\documentclass[runningheads]{llncs}
\usepackage{graphicx}
\usepackage{amsmath,amssymb} 
\usepackage{color}
\usepackage{epstopdf}
\usepackage{subfigure}
\usepackage{multirow}
\usepackage{array}
\usepackage{floatrow}
\newfloatcommand{capbtabbox}{table}[][\FBwidth]
\begin{document}
\title{CurriculumNet: Weakly Supervised Learning from Large-Scale Web Images}

%
\author{Sheng Guo \and
Weilin Huang\thanks{Weilin Huang is the corresponding author (e-mail:whuang@malong.com).} \and
Haozhi Zhang \and Chenfan Zhuang \and Dengke Dong  \and Matthew R. Scott \and Dinglong Huang}
\authorrunning{S. Guo \and W. Huang \and
H. Zhang \and C. Zhuang \and D. Dong  \and M. R. Scott \and D. Huang}
%

\institute{Malong Technologies, Shenzhen, China \\
Shenzhen Malong Artificial Intelligence Research Center, Shenzhen, China
\email{\{sheng,whuang,haozhang,fan,dongdk,mscott,dlong\}@malong.com}}
%
\maketitle              
\begin{abstract}
We present a simple yet efficient approach capable of training deep neural networks on large-scale weakly-supervised web images, which are crawled raw from the Internet by using text queries, without any human annotation. We develop a principled learning strategy by leveraging curriculum learning, with the goal of handling a massive amount of noisy labels and data imbalance effectively. We design a new learning curriculum by measuring the complexity of data using its distribution density in a feature space, and rank the complexity in an unsupervised manner. This allows for an efficient implementation of curriculum learning on large-scale web images, resulting in a high-performance CNN model, where the negative impact of noisy labels is reduced substantially. Importantly, we show by experiments that those images with highly noisy labels can surprisingly improve the generalization capability of the model, by serving as a manner of regularization. Our approaches obtain state-of-the-art performance on four benchmarks: WebVision, ImageNet, Clothing-1M and Food-101. With an ensemble of multiple models, we achieved a top-5 error rate of 5.2\% on the WebVision challenge  \cite{li2017webvision} for 1000-category classification. This result was the top performance by a wide margin, outperforming second place by a nearly 50\% relative error rate. Code and models are available at: https://github.com/MalongTech/CurriculumNet. 

\keywords{Curriculum learning \and  weakly supervised \and noisy data \and   large-scale \and web images}
\end{abstract}
\section{Introduction}
Deep convolutional networks have rapidly advanced numerous computer vision tasks, providing state-of-the-art performance on image classification \cite{HeK2016,SimonyanZ14a,SzegedyLJSRAEVR15,ioffe2015batch,Wang2017,Guo2017}, object detection \cite{Ren2015,redmon2016you,liu2016ssd,lin2017focal}, sematic segmentation \cite{long2015fully,hong2015decoupled,chen2016deeplab,he2017mask}, etc.  They produce strong visual features by training the networks in a fully-supervised manner using large-scale manually annotated datasets, such as ImageNet \cite{DengDSLL009}, MS-COCO \cite{lin2014microsoft} and  PASCAL VOC \cite{pascavoc}. Full and clean human annotations are of crucial importance to achieving a high-performance model, and better results can be reasonably expected if a larger dataset is provided with noise-free annotations. However, obtaining massive and clean annotations are extremely expensive and time-consuming, rendering the capability of deep models unscalable to the size of collected data. Furthermore, it is particularly hard to collect clean annotations for tasks where expert knowledge is required, and labels provided by different annotators are possibly inconsistent.

\begin{figure}[tb]
	\begin{center}
		\includegraphics[height=6.5cm,width=12cm]{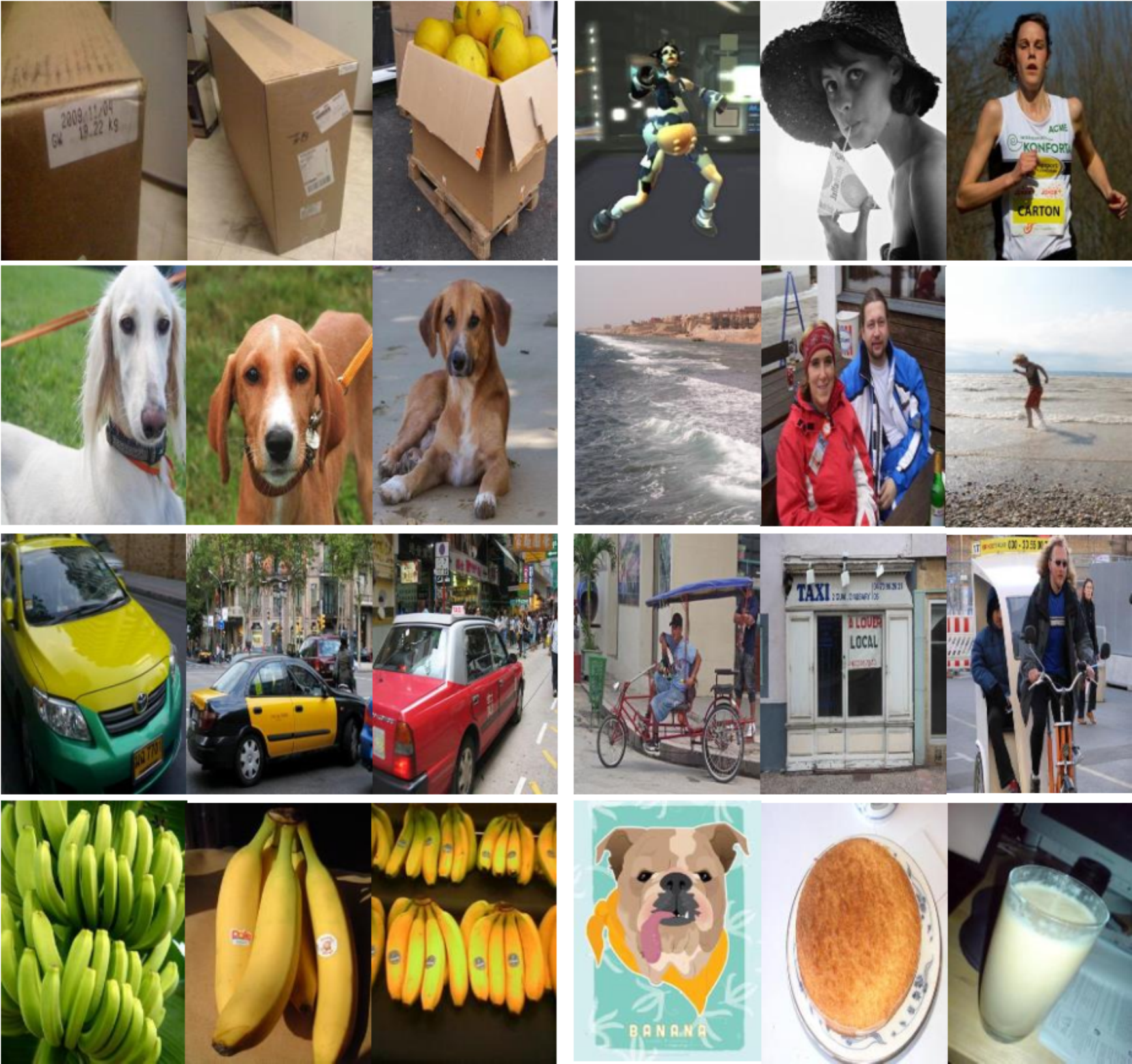}
	\end{center}
\vspace{-5mm}
	\caption{Image samples of the WebVision dataset \cite{li2017webvisiondata} from the categories of \emph{Carton}, \emph{Dog}, \emph{Taxi} and \emph{Banana}. The dataset was collected from the Internet by using text queries generated from the 1, 000 semantic concepts of the ImageNet benchmark \cite{DengDSLL009}. Each category includes a number of mislabeled images as shown on the right. }
	\label{fig:webvision_images}
\end{figure}

An alternative solution is to use the web as a source of data and supervision, where a large amount of web images can be collected automatically from the Internet by using input queries, such as text information.
These queries can be considered as natural annotations of the images, providing weak supervision of the collected data, which is a cheap way to increase the scale of the dataset near-infinitely. However, such annotations are highly unreliable, and often include a massive amount of noisy labels. Past work has shown that these noisy labels could significantly affect the performance of deep neural networks on image classification \cite{xiao2015learning}. To address this problem, recent approaches have been developed by proposing robust algorithms against noisy labels \cite{rolnick2017deep}. Another solution is to develop noise-cleaning methods that aim to remove or correct the mislabelled examples in training data \cite{sukhbaatar2014learning}. However, the noise-cleansing methods often suffer from the main difficulty in distinguishing mislabeled samples from hard samples, which are critical to improving model capability. Besides, semi-supervised methods have also been introduced by using a small subset of manually-labeled images, and then the models trained on this subset are generalized to a larger dataset with unlabelled or weakly-labelled data \cite{Veit2017}.  Unlike these approaches, we do not aim to propose a noise-cleaning, noise-robust or semi-supervised algorithm. Instead, we investigate improving model capability of standard neural networks by introducing a new training strategy.

In this work, we study the problem of learning convolutional networks from large-scale images with a massive amount of noisy labels, such as the WebVision challenge \cite{li2017webvision},  which is a  1000-category image classification task having the same categories as ImageNet \cite{DengDSLL009}. The labels are provided by simply using the queries text generated from the 1,000 semantic concepts of ImageNet \cite{DengDSLL009}, \emph{without any manual annotation}. Several image samples are presented in Fig. \ref{fig:webvision_images}. Our goal is to provide a solution able to handle massive noisy labels and data imbalance effectively. We design a series of experiments to investigate the impact of noisy labels on the performance of deep networks, when the amount of training images is sufficiently large. We develop a simple but surprisingly efficient training strategy that allows for improving model generalization and overall capability of the standard deep networks, by leveraging highly noisy labels. We observe that training a CNN from scratch using both clean and noisy data is more effective than just using the clean one. The contributions of this work are three-fold:
\begin{itemize}
  \item[--] We propose CurriculumNet by developing an efficient learning strategy with curriculum learning. This allows us to train high-performance CNN models from large-scale web images with massive noisy labels, which are obtained without any human annotation.

   \item[--] We design a new learning curriculum by ranking data complexity using distribution density in an unsupervised manner. This allows for an efficient implementation of curriculum learning tailored for this task, by directly exploring highly noisy labels.

   \item[--] We conduct extensive experiments on a number of benchmarks, including WebVision \cite{li2017webvisiondata}, ImageNet \cite{DengDSLL009}, Clothing1M \cite{xiao2015learning} and Food101 \cite{bossard2014food}, where the proposed CurriculumNet obtains state-of-the-art performance. The CurriculumNet, with an ensemble of multiple models, archived the top performance with a top-5 error rate of 5.2\%, on the WebVision Challenge at CVPR 2017, outperforming the other results by a large margin.
\end{itemize}

\section{Related work}
We give a brief review on recent studies developed for dealing with noisy annotations on image classification. For a comprehensive overview of label noise taxonomy and noise
robust algorithms we refer to \cite{frenay2014classification}.

Recent approaches to learn from noisy web data can be roughly classified into two categories. (1) Methods aim to directly learn from noisy labels. This group of approaches mainly focus on noise-robust algorithms \cite{larsen1998design,xiao2015learning,Patrini2017}, and label cleansing methods which aim to remove or correct mislabeled data \cite{Brodley1999,jiang2017mentornet}. However, they generally suffer from the main challenge of identifying mislabeled samples from hard training samples, which are crucial to improve model capability.
(2) Semi-supervised learning approaches have also been developed to handle these shortcomings, by combining the noisy labels with a small set of clean labels \cite{Zhu2005,Fergus2009,Chen2013}.  A transfer learning approach solves the label noise by transferring correctness of labels to other classes \cite{Lee2017}. The models trained on this subset are generalized to a larger dataset with unlabelled or weakly-labelled data \cite{Veit2017}.  Unlike these approaches, we do not propose a noise-cleansing or noise-robust or semi-supervised algorithm. Instead, we investigate improving model capability of the standard neural networks, by introducing a new training strategy that alleviates negative impact of the noisy labels.

Convolutional neural networks have recently been applied to training a robust model with noisy data \cite{xiao2015learning,rolnick2017deep,Patrini2017,Lee2017,jiang2017mentornet}.  Xiao \emph{et al.} \cite{xiao2015learning} introduced a general framework to train CNNs with a limited amount of human annotation, together with noisy data sets containing millions of images.   A  behavior of CNNs on the training set with highly noisy labels was studied in \cite{rolnick2017deep}.  MentorNet \cite{jiang2017mentornet} improved the performance of CNNs trained on noisy data, by learning an additional network that weights the training examples. Our method differs from these approaches by directly considering the mislabelled samples in our training process, and we show by experiments that with an efficient training scheme, a standard deep network is strongly robust against the highly noisy labels.

Our work is closely related to the work of \cite{Misra2016}, which is able to model noise arising from missing, but visually present labels. The method in \cite{Misra2016} is conditioned on the input image, and was designed
for multiple labels per image. It does not take advantage of cleaning labels, and the focus is on missing labels, while our approach works reliably on the highly noisy labels, without any cleaned (manual annotation). Our learning curriculum is designed in a completely unsupervised manner.

\begin{figure*}[tb]
	\begin{center}
		\includegraphics[width=12cm]{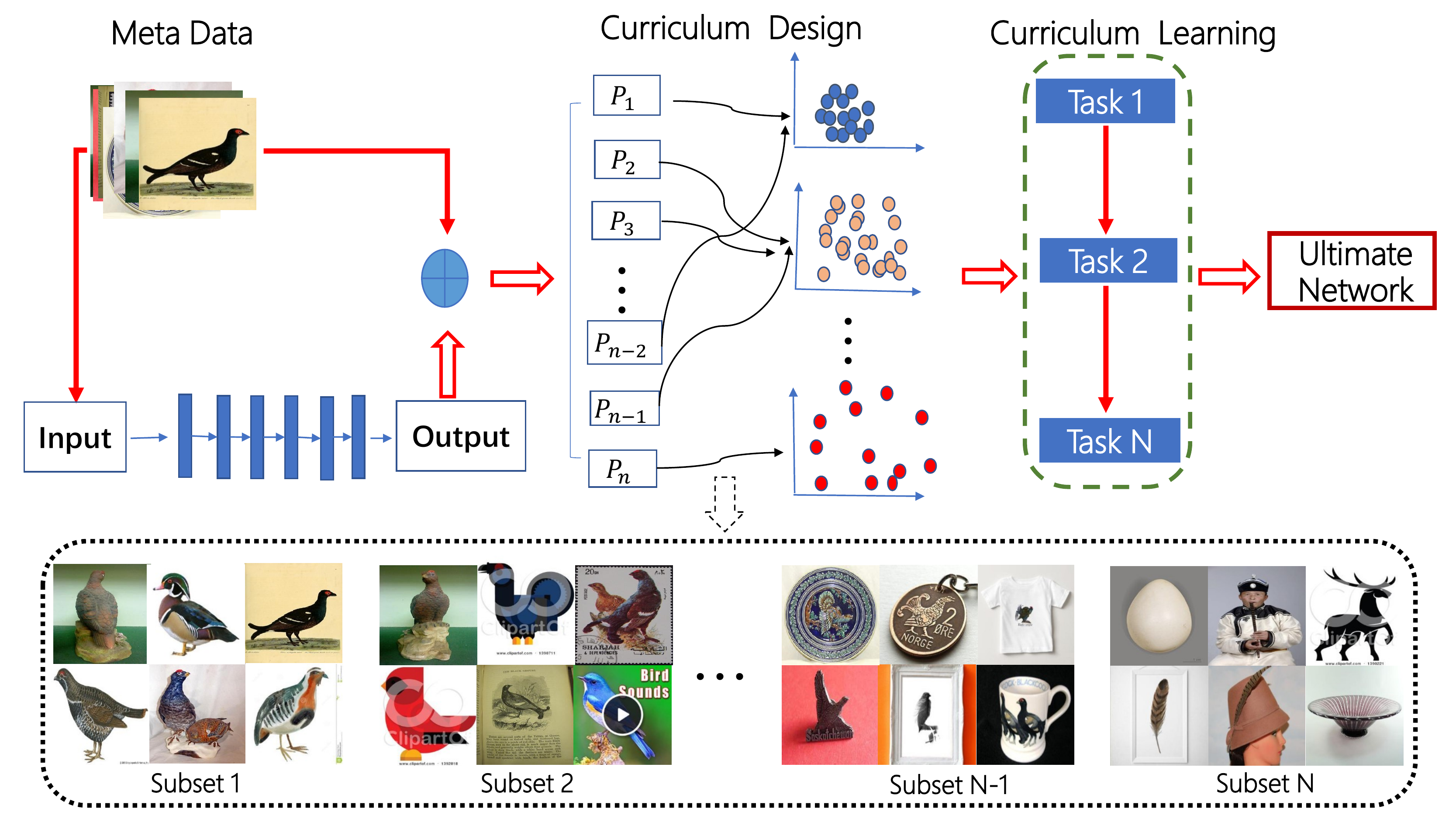}
	\end{center}
\vspace{-5mm}
	\caption{Pipeline of the proposed CurriculumNet. The training process includes three main steps： initial features generation, curriculum design and curriculum learning.}
	\label{fig:main}
\end{figure*}

\section{Methodology}

In this section, we present details of the proposed CurriculumNet motivated by human learning, in which the model starts from learning easier aspects of a concept, and then gradually includes more complicated tasks into the  learning process \cite{bengio2009curriculum}. We introduce a new method to design a learning curriculum in an unsupervised manner. Then CNNs are trained by following the designed curriculum, where the amount of noisy labels is increased gradually.

\subsection{Overview}

 Pipeline of CurriculumNet is described in Fig. \ref{fig:main}. It contains three main steps: (i) initial features generation, (ii) curriculum design and (iii) curriculum learning. First, we use all training data to learn an initial model which is then applied to computing a deep representation (e.g., fully-convolutional (fc) features) from each image in the training set. Second, the initial model aims to roughly map all training images into a feature space where the underlying structure and relationship of the images in each category can be discovered, providing an efficient approach that defines the complexity of the images. We explore the defined complexity to design a learning curriculum where all images in each category are split into a number of subsets ordered by complexity.
 Third, based on the designed curriculum, we employ curriculum learning which starts training CNNs from an easy subset which combines the easy subsets over all categories. It is assumed to have more clean images with correct labels in the easy subset. Then the model capability is improved gradually by continuously adding the data with increasing complexity into the training process.

\subsection{Curriculum Design}

Curriculum learning was originally proposed in \cite{bengio2009curriculum}. It was recently applied to dealing with noise and outliers. One of the main issues to deliver advances of this learning idea is to design an efficient learning curriculum that is specific for our task. The designed curriculum should be able to discover meaningful underlying local structure of the large-scale noisy data in a particular feature space, and our goal is to design a learning curriculum able to rank the training images from easy to complex  in an unsupervised manner. We apply a density based clustering algorithm that measures the complexity of training samples using data distribution density. Unlike previous approaches which were developed to handle noisy labels in small-scale or moderate-scale datasets, we design a new learning curriculum that allows our training strategy with a standard CNN to work practically on large-scale datasets, e.g., the WebVision database which contains over 2,400,000 web images with massive noisy labels.

Specifically, we aim to split the whole training set into a number of subsets, which are ranked from an easy subset having clean images with more reliable labels, to a more complex subset containing massive noisy labels. Inspired by recent clustering algorithms described in \cite{rodriguez2014clustering},  we conduct the following procedures \emph{in each category}. First, we train an initial model from the whole training set by using an Inception\_v2 architecture \cite{ioffe2015batch}. Then all images in each category are projected into a deep feature space, by using the \emph{fc}-layer features of the initial model, $P_i \rightarrow f(P_i)$ for each image $P_i$. Then we calculate a Euclidean distance matrix $D \subseteq \mathbb{R}^{n\times n}$ as,
\begin{equation}
D_{ij}=\|f(P_i )-f(P_j)\|^2
\end{equation}
where $n$ is the number of images in current category, and $D_{ij}$ indicates a similarity value between $P_i$ and $P_j$( A smaller $D_{ij}$ means higher similarity between $P_i$ and $P_j$ ).

We first calculate a local density $(\rho_i)$ for each image:
\begin{equation}
\rho_i=\sum_{j}X(D_{ij}-d_c)
\end{equation}
where
$$X(d)=
\begin{cases}
1& d<0\\
0& \text{other}
\end{cases}$$
where $d_c$ is determined by sorting $n^2$ distances in $D \subseteq \mathbb{R}^{n\times n}$ from small values to large ones, and select a number which is ranked at $k\%$. This result is insensitive to the value of $k$ between 50 and 70, and we empirically set $k=60$ in all our experiments. $\rho_i$ is the number of samples whose distances to $i$ is smaller than $d_c$. It is natural to assume that a group of clean images with correct labels often have relatively similar visual appearance, and these images are projected closely to each other, leading to a large value of local density. By contrast, noisy images often have a significant visual diversity, resulting in a sparse distribution with a smaller value of the density.

Then we define a distance $(\delta_i)$ for each image:
\begin{equation}
\delta_i=
\begin{cases}
min_{j:\rho_j>\rho_i}(D_{ij})& if \; \exists j \; s.t. \; \rho_j>\rho_i \\
max(D_{ij}) & \text{otherwise}
\end{cases}
\end{equation}

If there exists an image $I_j$ having $ \rho_j>\rho_i$, $\delta_i$ is $D_{i\hat{j}}$ where $\hat{j}$ is the sample nearest to $i$ among the data. Otherwise, if $\delta_i$ is the largest one among all density, $\rho_j$ is the distance between $i$ and the data point which is farthest from $i$. Then a data point with the highest local density has the maximum value of $\delta$, and is selected as cluster center for this category.

\begin{figure}[tb]
	\begin{center}
		\includegraphics[height=4cm,width=6cm]{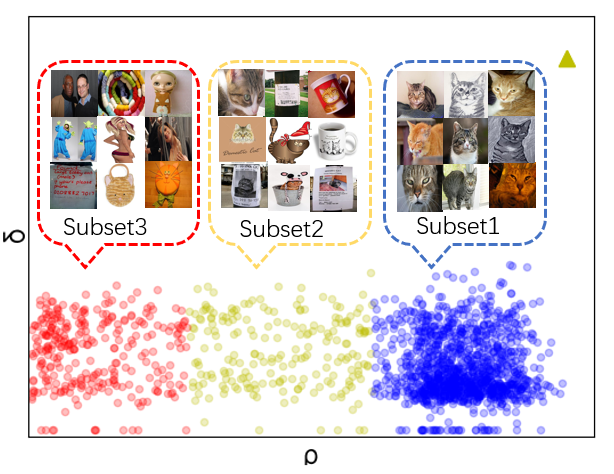}
		\includegraphics[height=4cm,width=6cm]{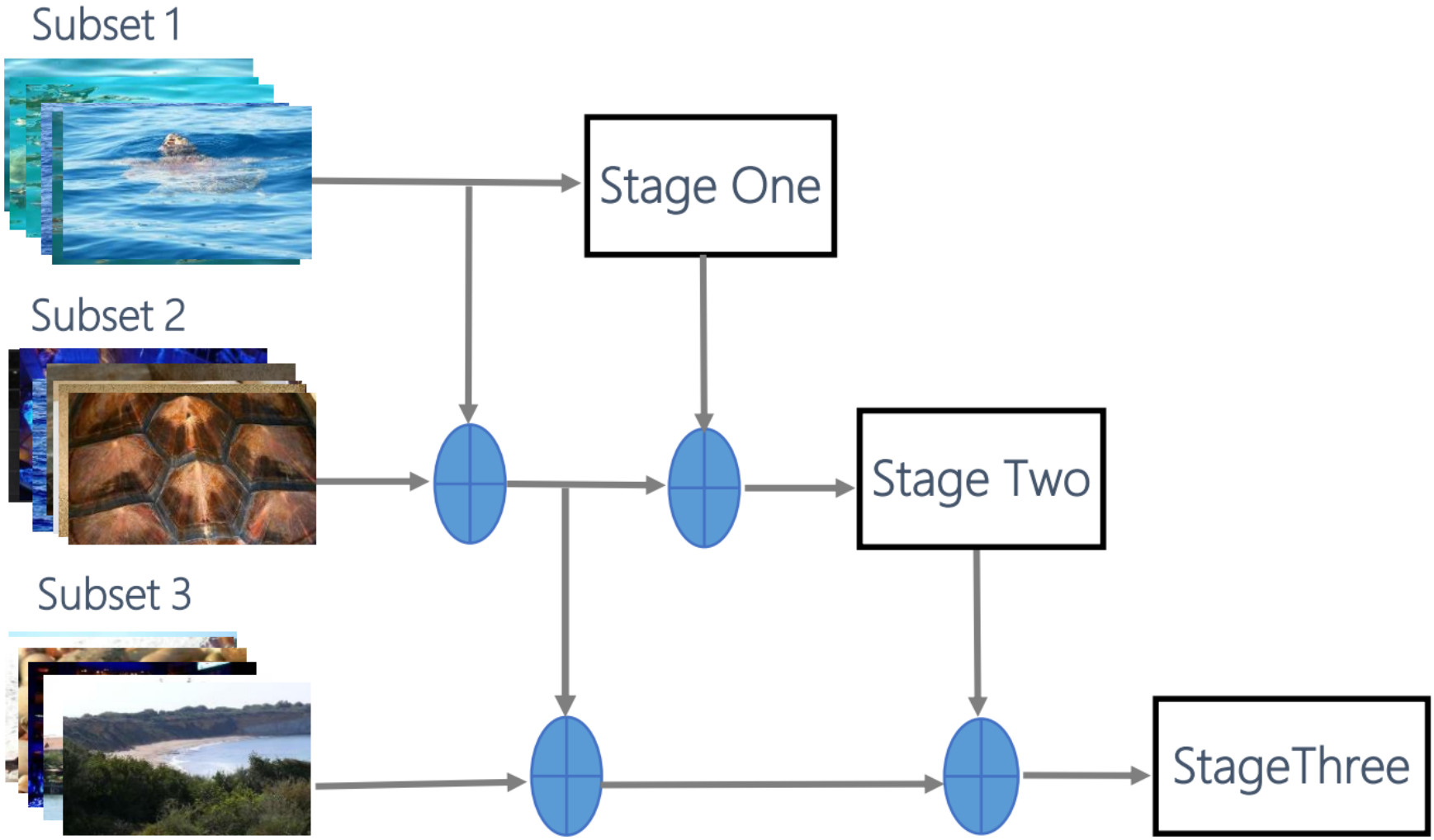}
	\end{center}
	\caption{\underline{Left}: the sample of the cat category with three subsets. \underline{Right}: learning process with designed curriculum.}
	\label{fig:curriculum_learning}
\end{figure}

As we have computed a cluster center for the category, a closer data point to the cluster center, has a higher confidence to have a correct label.
Therefore, we simply proceed with the k-mean algorithm to divide data points into a number of clusters, according to their distances to the cluster center, $D_{cj}$, where $c$ is the cluster center. Fig. \ref{fig:curriculum_learning} (left) is an $\delta- \rho$ figure for all images in the category of cat from the WebVision dataset.

We generate three clusters in each category, and simply use the images within each cluster as a data subset. As each cluster has a density value measuring data distribution within it, and relationship between different clusters. This provides a natural way to define the complexity of the subsets, giving a simple rule for designing a learning curriculum. A subset with a high density value means all images are close to each other in feature space, suggesting that these images have a strong similarity. We define this subset as a \emph{clean} one, by assuming most of the labels are correct. The subset with a small density value means the images have a large diversity in visual appearance, which may include more irrelevant images with incorrect labels. This subset is considered as \emph{noisy} data. Therefore, we generate a number of subsets in each category, arranged from clean, noisy, to highly noisy ones, which are ordered with increasing complexity. Each category has the same number of  subsets, and we combine them over all categories, which form our final learning curriculum that implements training sequentially on the clean, noisy and highly noisy subsets.  Fig. \ref{fig:curriculum_learning} (left) show data distribution of the three subsets in the category of ``cat" from the WebVision dataset, with a number of sample images. As seen, images from the clean subset have very close visual appearance, while the highly noisy subset contains a number of random images which are completely different from those in the clean subset.

\subsection{Curriculum Learning}
The Learning process is performed by following the nature of the underlying data structure. That is, the designed curriculum is able to discover the underlying data structure based on visual appearance, in an unsupervised manner. We design a learning strategy which relies on intuition - tasks are ordered by increasing difficulty, and training is proceeded sequentially from easier tasks to harder ones. We develop a multi-stage learning process that trains a standard neural network more efficiently with the enhanced capability for handling massive noisy labels.

Training details are described in Fig. \ref{fig:curriculum_learning} (right), where a convolutional model is trained through three stages by continuously mixing training subsets from the clean subset to the highly noisy one. Firstly, a standard convolutional architecture, such as Inception\_v2 \cite{ioffe2015batch}, is used. The model is trained by only using the clean data, where images within each category have close visual appearance. This allows the model to learn basic but clear visual information from each category, serving as the fundamental features for the following process. Secondly, when  the model trained in the first stage converges, we continue the learning process by adding the noisy data, where images have more significant visual diversity, allowing the model to learn more meaningful and discriminative features from harder samples. Although the noisy data may include incorrect labels, it roughly preserves the main structure of the data, and thus leads to performance improvement. Thirdly, the model is further trained by adding the highly noisy data which contains a large number of visually irrelevant images with incorrect labels. The deep features learned by following the first two-stage curriculum are able to capture the main underlying structure of the data. We observe that the highly noisy data added in the last stage does not impact negatively to the learned data structure. By contrast, it improves the generalization capability of the model, and allows the model to avoid over-fitting over the clean data, by providing a manner of regularization. A final model is obtained when the training  converges in the last stage, where the three subsets are all combined. In addition, when samples from different subsets are combined in the second and third stages, we set different loss weights to the training samples of different subsets as 1, 0.5 and 0.5 for the clean, noisy and highly noisy subsets, respectively.

\subsection{Implementation Details}

\noindent\textbf{Training Details}: The scale of WebVision data \cite{li2017webvisiondata} is significantly larger than  that of ImageNet \cite{DengDSLL009}, it is important to consider the computational cost when extensive experiments are conducted in evaluation and comparisons. In our experiments, we employ the inception architecture with batch normalization (bn-inception) \cite{ioffe2015batch} as our standard architecture.  The bn-inception model is trained by adopting the proposed density-ranking curriculum leaning. The network weights are optimized with mini-batch stochastic gradient decent (SGD), where the batch size is set to 256, and Root Mean Square Propagation (RMSprop) algorithm \cite{ioffe2015batch} is adopted. The learning rate starts from 0.1, and decreases by a factor of 10 at the iterations of $30\times10^4, 50\times 10^4, 60\times 10^4, 65\times10^4, 70\times10^4$.
The whole training process stop at $70\times 10^4$ iterations. To reduce the risk of over-fitting, we use common data augmentation technologies which include random cropping , scale jittering, and ratio jittering. We also add a dropout operation with a ratio of 0.2 after the global pooling layer. \\

\noindent \textbf{Selective Data Balance}: By comparing with ImageNet, another challenge of the WebVision data \cite{li2017webvision} is that the training images in different categories are highly unbalanced. For example, a large-scale category can have over 10,000 images, while a small-scale category only contains  less than 400 images. CNN models, directly trained with random sampling on such unbalanced classes, will have a bias towards the large categories. To alleviate this problem, we develop a two-level data balance approach: subset-level balance and category-level balance. In the subset-level balance, training samples are selected in each min-batch as follows: (256, 0, 0), (128, 128, 0) and (128, 64, 64) for stage 1-3, respectively. For the category-level balance, in each mini-batch, we first random select 256 (in stage 1) or 128 (in stage 2 and 3) categories from the 1000 classes, and then we randomly select only one sample from each selected category. Notice that the category-level balance is only implemented on the clean subset. The performance was dropped down when we applied it to the noisy or highly noisy subset. Because we randomly collect a single sample from each category in the category-level balance, it is possible to obtain a single but completely irrelevant sample from the noisy or highly noisy subset, which would negatively affect the training.\\

\noindent \textbf{Multi-scale convolutional kernels:} We also apply multi-scale convolutional kernels in the first convolutional layer, with three different kernel sizes: $5 \times 5$, $7 \times 7$ and $9 \times 9$. Then we concatenate three convolutional maps generated by three types of filters, which form the final feature maps of the first convolutional layer. The multi-scale filters enhance the low-level features in the first layer, leading to about 0.5\% performance improvements on top-5 errors on the WebVision data.


\section{Experimental Results and Comparisons}

The proposed CurriculumNet is evaluated on four benchmarks:  WebVision \cite{li2017webvisiondata}, ImageNet \cite{DengDSLL009}, Clothing1M \cite{xiao2015learning} and Food101 \cite{bossard2014food}. Particularly, we investigate the learning capability on large-scale web images without human annotation. 

\subsection{Datasets}

\noindent \textbf{WebVision} dataset \cite{li2017webvisiondata} is an  object-centric dataset, and is larger than  ImageNet \cite{DengDSLL009}  for object recognition  and classification. The images are crawled from both Flickr and Google images search, by using queries generated from the 1, 000 semantic concepts of the ILSVRC 2012. Meta information along with those web images (e.g., title, description, tags, etc.) are also crawled. The dataset for the WebVision 2017 contains 1,000 object categories  (the same with  the ImageNet). The training data contains 2,439,574 images in total, but without any human annotation. It includes massive noisy labels, as shown in Fig. \ref{fig:webvision_images}. There are 50,000  manually-labeled images are used as validation set, and another 50,000 manually-labeled images for testing. The evaluation measure is based on top-5 error, where each algorithm provides a list of at most 5 object categories to match the ground truth. \\

\noindent \textbf{Clothing1M} dataset \cite{xiao2015learning} is a large-scale fashion dataset, which includes 14 clothes categories. It contains 1 million noisy labeled images and 74,000 manually annotated images. We call the annotated images the clean set, which is divided into training data, validation data and testing data, with numbers of 50,000, 14,000, and 10,000 images, respectively. There are some images that overlap between the clean set and the noisy set. The dataset was designed for learning robust models from noisy data without human supervision.\\

\noindent \textbf{Food-101} dataset \cite{bossard2014food} is a standard benchmark to evaluate recognition accuracy of food visuals. It contains 101 classes, with 101,000 real-world food images in total. The numbers of training and testing images are 750 and 250 per category, respectively. This is a clean dataset with full manual annotations provided. To conduct experiments with noisy data, we manually add 20\% noisy images into the training set, which are randomly collected from the training set of ImageNet \cite{DengDSLL009}, and each image is randomly assigned a label from 101 categories from the Food-101.

\subsection{Experiments and Comparisons}
We conducted extensive experiments to evaluate the efficiency of the proposed approaches. We compare various training schemes by using the BN-Inception. 

\subsubsection{On training strategy.}
 We evaluate four different training strategies by using a standard Inception\_v2 architecture, resulting in four models, which are described as follows.
\begin{itemize}
\item[--] \textbf{Model-A}: the model is trained by directly using the whole training set.
\item[--] \textbf{Model-B}: the model is trained by only using the clean subset.
\item[--] \textbf{Model-C}: the model is trained by using the proposed learning strategy, with a 2-subset curriculum: clean and noisy subsets.
\item[--] \textbf{Model-D}: the model is trained by using the proposed learning strategy, with a 3-subset curriculum: clean, noisy and highly noisy subsets.
\end{itemize}

\begin{figure*}[tb]
\centering
\subfigure{\includegraphics[width=6cm]{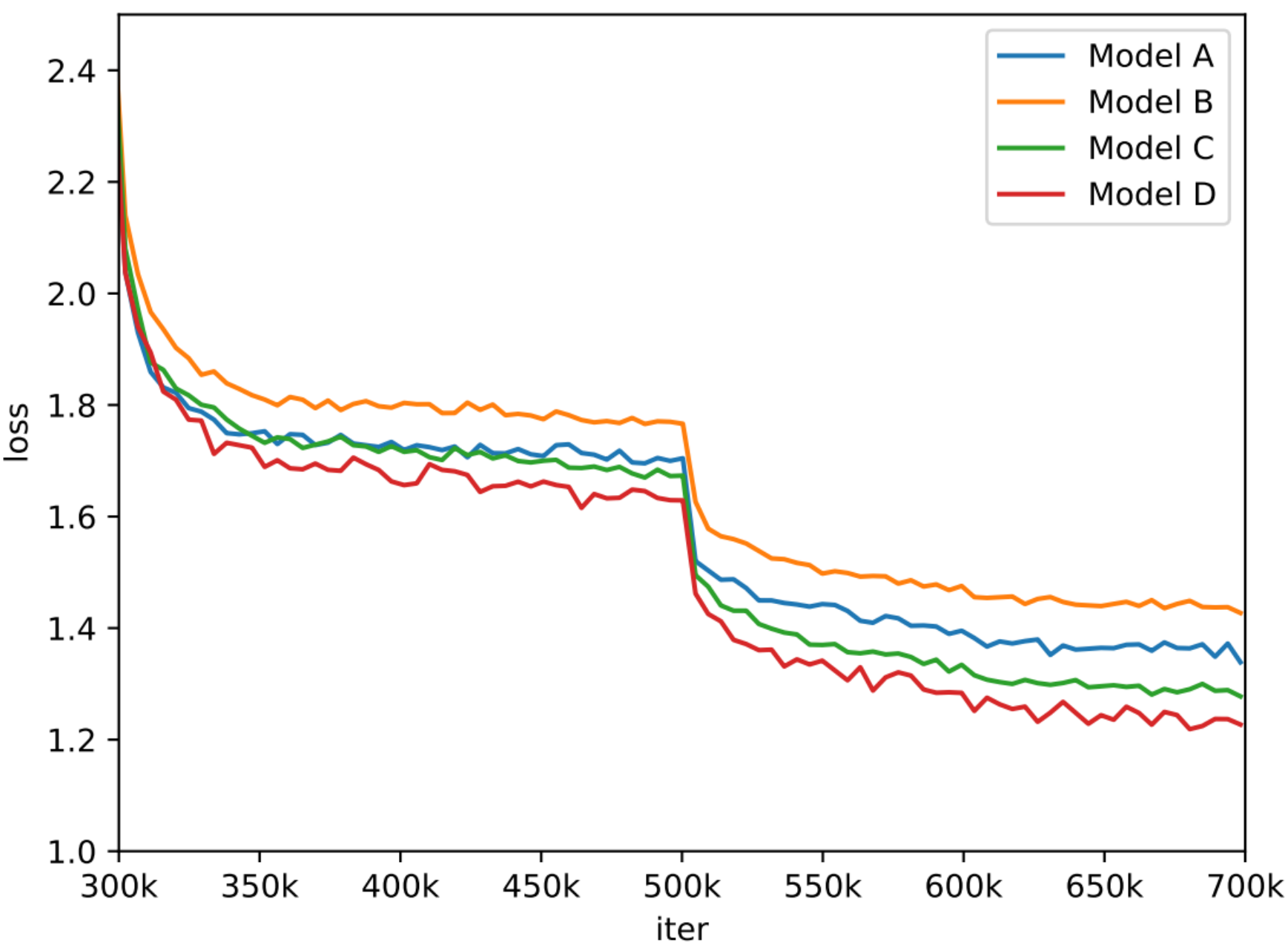}}
\subfigure{\includegraphics[width=6cm]{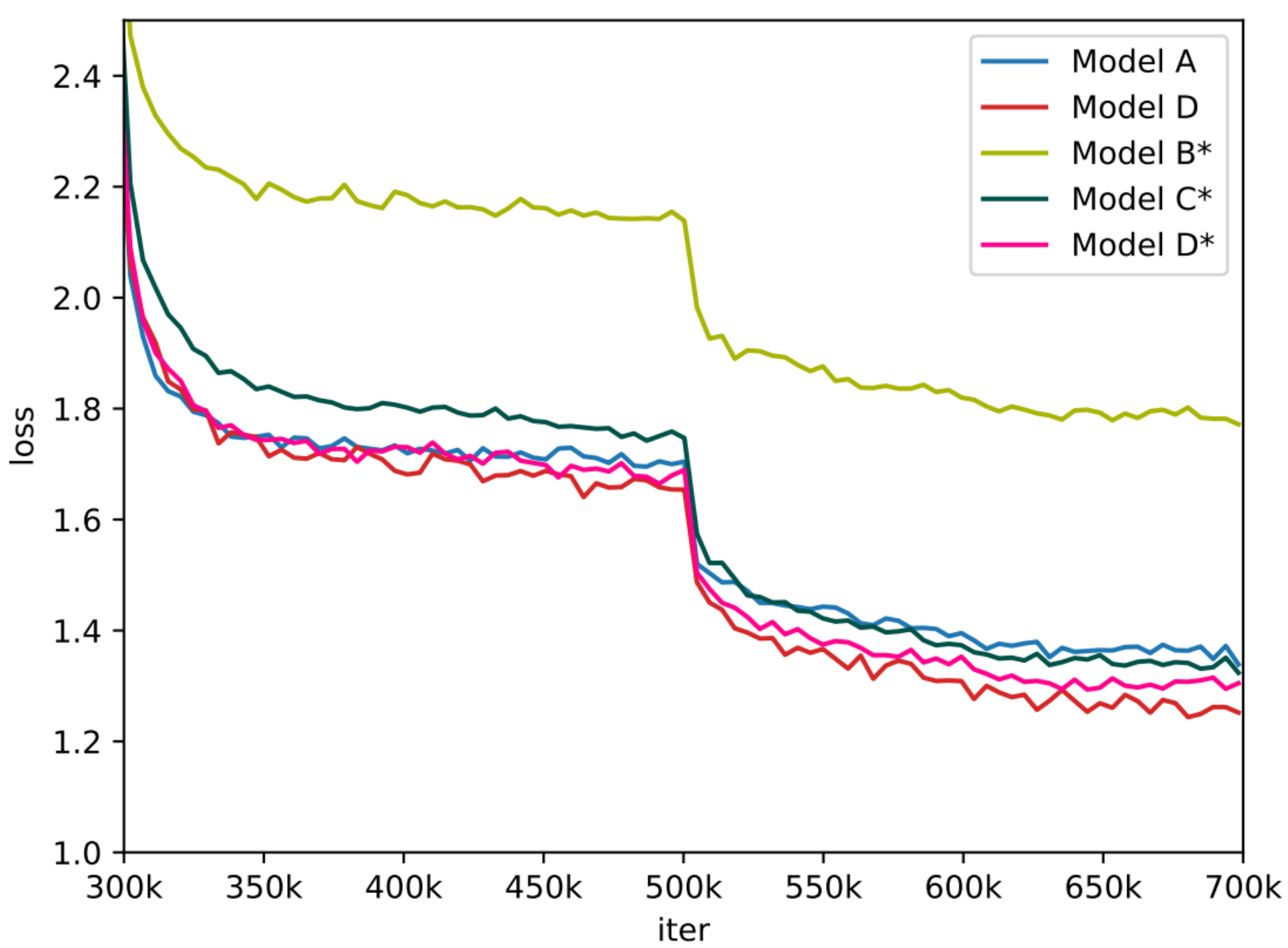}}
\vspace{-5mm}
\caption{Testing loss of four different models with BN-Inception architecture, (left) Density-based curriculum, and (right) K-mean based curriculum.}
\label{fig:testloss}
\end{figure*}

Test loss of four models (on the validation set of WebVision) are compared in Fig. \ref{fig:testloss}, where the proposed CurriculumNet with a 2-subset curriculum and a 3-subset curriculum (Model-C and Model-D) have better convergence rates.
Top 1 and Top 5 results of four models on the validation set of WebVision are reported in Table \ref{tbl:test_model}. The results are mainly consistent with the test loss presented in Fig. \ref{fig:testloss}. The proposed method, with 3-subset curriculum learning, significantly outperforms the model trained on all data, with improvements of $30.16\% \rightarrow 27.91\%$ and $12.43\% \rightarrow 10.82\%$ on Top 1 and Top 5 errors, respectively. These improvements are significant on such a large-scale challenge. Consistent improvements are obtained on the validation set of ImageNet, where the models were trained on the WebVision data. In all 1000 categories, our approaches lead to performance improvements on 668 categories, while only 195 categories reduced their Top 5 results, and the results of remaining 137 categories were unchanged.




\begin{table}[tp]
\begin{center}
\caption{Top-1 and Top-5 errors (\%) of four different models with BN-Inception architecture on validation set. The models are trained on  WebVision training set and tested on the WebVision and ILSVRC validation sets under various models.}
\label{tbl:test_model}
\vspace{-2mm}
\begin{tabular}{p{1.8cm}|p{1.6cm}<{\centering}|p{1.6cm}<{\centering}|p{1.6cm}<{\centering}|p{1.6cm}<{\centering}}
  \hline
 \multirow{2}{*}{Mothed}& \multicolumn{2}{c}{\textbf{ WebVision}}&  \multicolumn{2}{c}{\textbf{ImageNet}} \\  \cline{2-5}
&Top-1&Top-5&Top-1&Top-5 \\
\hline
  Model-A & 30.16 &  12.43 &36.00&16.20 \\
  \hline
  Model-B & 30.28& 12.98&37.09&16.42 \\
    \hline
  Model-C  & 28.44 & 11.38 &35.66&15.24 \\
  \hline
  Model-D  & {\bf27.91} &{\bf 10.82} &{\bf 35.24}& {\bf 15.11}  \\
  \hline
\end{tabular}

\end{center}
\end{table}

%

\subsubsection{On highly noisy data or training labels.}
We further investigate the impact of highly noisy data to the proposed learning strategy. We used different percentages of data from the highly noisy subset for 3-subset curriculum learning, ranging from 0\% to 100\%. Results are reported in Table \ref{tbl:test_model_noisy}. As shown, the best results on both Top 1 and Top 5 are achieved at 50\% of the highly noisy data. This suggests that, by using the proposed training method, even the highly noisy data can improve model generalization capability, by increasing the amount of the training data with more significant diversity, demonstrating the efficiency of the proposed approach. Increasing the amount of highly noisy data further did not improvement the performance, but with very limited negative affect.

To provide more insights and give deeper analysis on the impact of label noise, we applied the most recent ImageNet-trained SEnet \cite{hu2017squeeze} (which has a Top 5 error of 4.47\%  on ImageNet) to classify all images from the training set of the WebVision data. \emph{We assume the output label of each image by SEnet is correct, and compute the rate of correct labels in each category}. We observed that the average noise rate over the whole training set of the WebVision data is high to 52\% (Top 1), indicating that a large amount of incorrect labels is included. We further compute the average noise rates for three subsets of the designed learning curriculum, which are 65\%, 39\% and 15\%, respectively. These numbers are consistent with the increasing complexity of the three subsets, and suggest that most of the images in the third subset are highly noisy.

\begin{figure*}[tb]
\setlength{\abovecaptionskip}{-5pt}
\setlength{\belowcaptionskip}{-9pt}
\centering
\subfigure{\includegraphics[width=6cm]{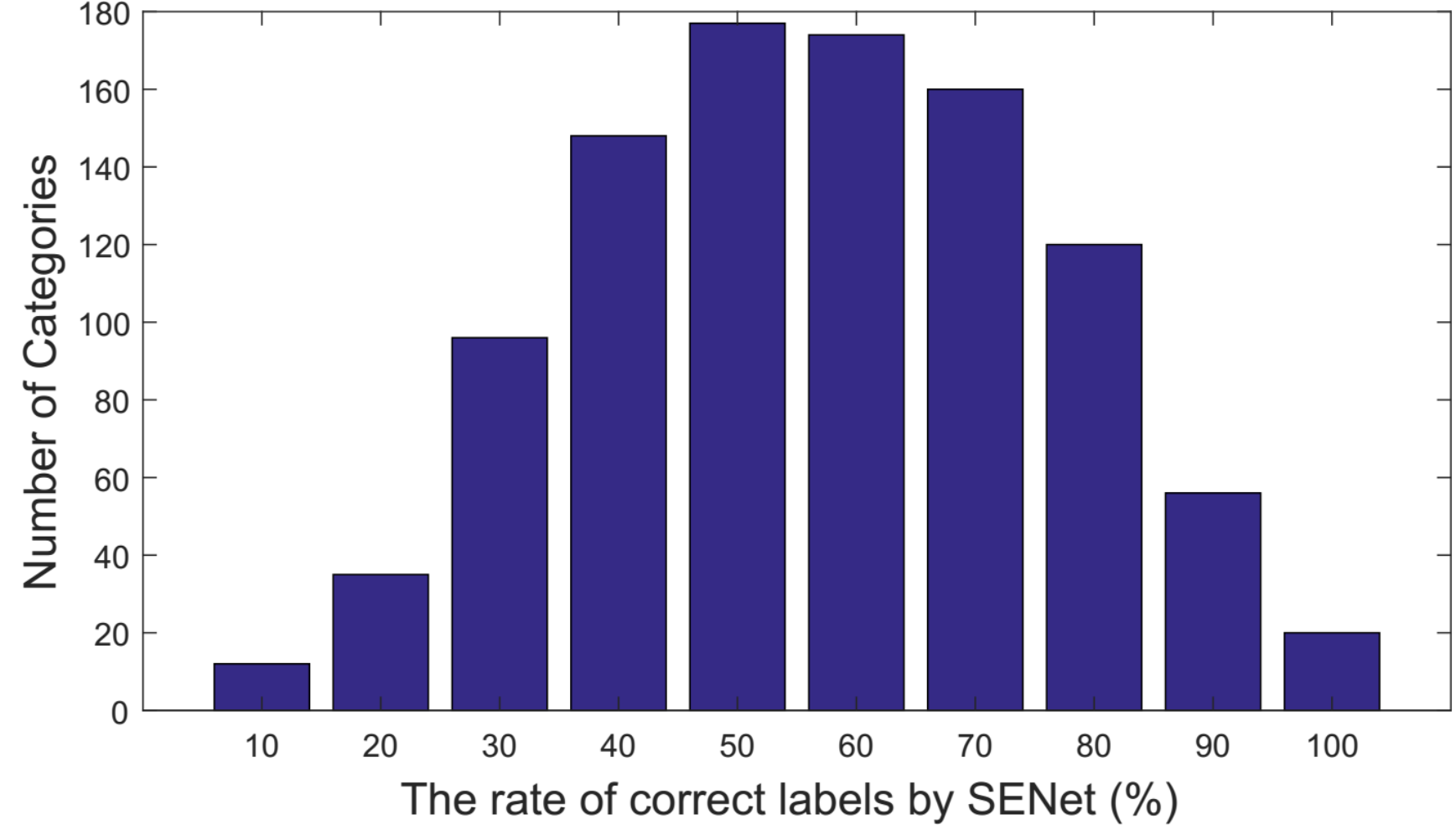}}
\subfigure{\includegraphics[width=6cm]{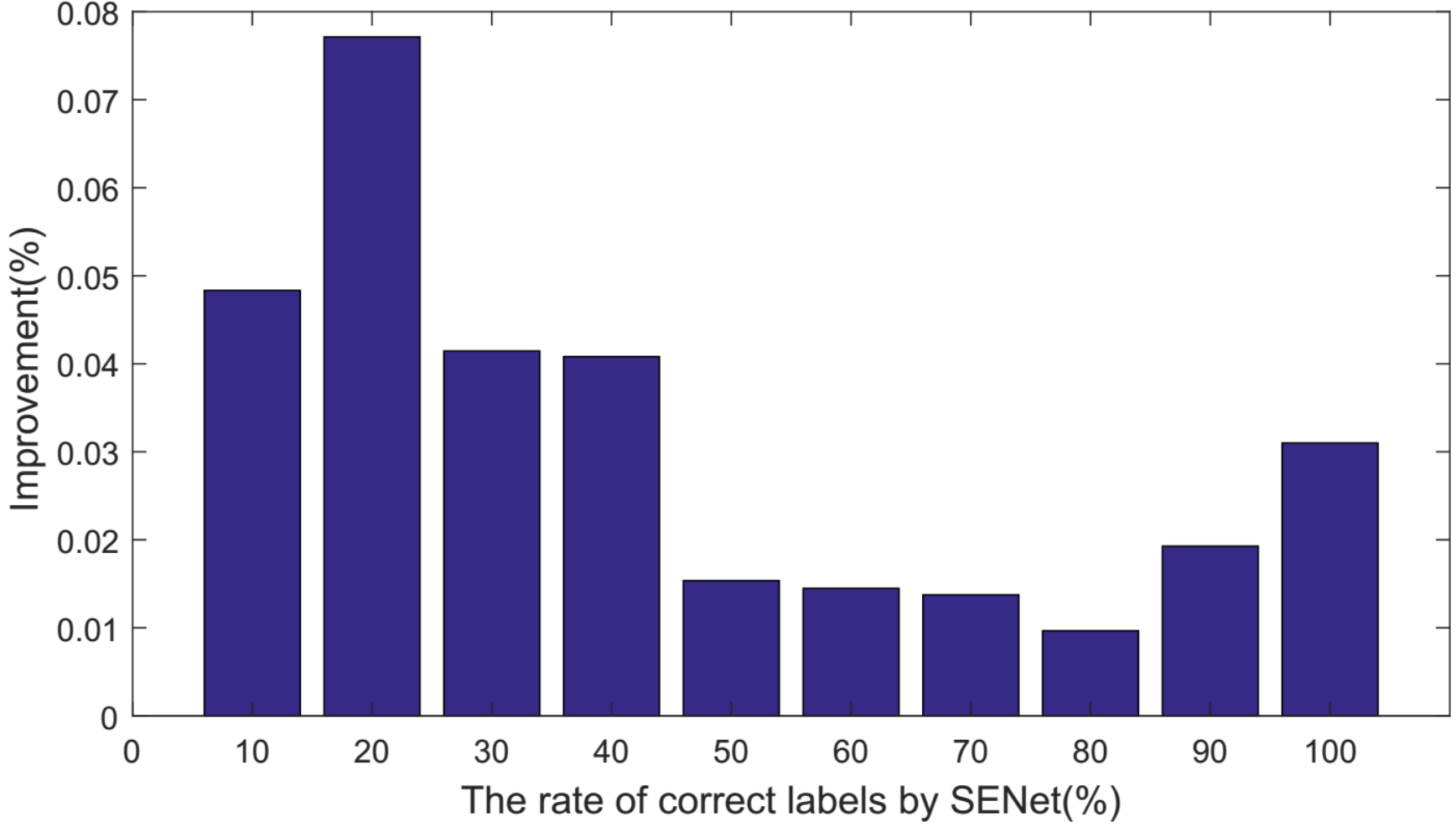}}
\vspace{-5mm}
\caption{Numbers of categories (left), and performance improvements (right) in 10 different rate intervals of the training labels.}
\label{fig:noiseRate}
\end{figure*}

We calculate the number of categories in 10 different intervals of the correct rates of the training labels, as shown in Fig. \ref{fig:noiseRate} (left). There are 12 categories having a correct rate that is lower than 10\%. We further compute the average performance gain in each interval, as show in Fig. \ref{fig:noiseRate} (right). We found that the categories with lower correct rates (e.g., $<40\%$) have larger performance gains ($>4\%$), and the most significant improvement happens in the interval of 10\%-20\%, which has an improvement of 7.7\%.



\subsubsection{On different clustering algorithms.}
The proposed clustering based curriculum learning can generalize well to other clustering algorithms. We verify it by comparing our density based curriculum design with K-means based clustering on the proposed 3-subset CurriculumNet. As shown in Fig. \ref{fig:testloss} (right), the Model-B* which is trained using the clean subset by K-means has a significantly lower performance, which means that training without the proposed curriculum learning is highly sensitive to the quality. By adopting the proposed method, Model-D* significantly improves the performance, from 16.6\% to 11.5\% (Top 5), which is comparable to Model-D. These results demonstrate the strong robustness of the proposed CurriculumNet, allowing for various qualities of the data generated by different algorithms.



\subsubsection{Final results on the WebVision challenge.}
We further evaluate the performance of CurriculumNet (Model-D) by using various network architectures, including Inception\_v2 \cite{ioffe2015batch}, Inception\_v3 \cite{szegedy2016rethinking}, Inception\_v4 \cite{szegedy2017inception} and Inception\_resnet\_v2 \cite{szegedy2017inception}. Results are reported in Table \ref{tbl:test_model_networks}.  As can be found, the Inception\_v3 outperforms the Inception\_v2 substantially, from  10.82\% to 7.88\% on the Top 5, while a more complicated model, such as Inception\_v4  and Inception\_resnet\_v2, only has similar performance with a marginal performance gain obtained.

Our final results were obtained with ensemble of six models. We had the best performance at a Top 5 error of 5.2\% on the WebVision challenge 2017\cite{li2017webvision}. It outperforms the 2nd one by a margin of about 2.5\%, which is about 50\% relative error, and thus is significant for this challenging task. The 5.2\% Top 5 error is also comparable to human performance on the ImageNet, but our method obtained this result by using weakly-supervised training data without any human annotation.

\begin{table*} [tp]
\begin{floatrow}
\capbtabbox{
\begin{tabular}{c|p{1.3cm}<{\centering}|p{1.3cm}<{\centering}}
  \hline
  Noise data(\%) & Top1 & Top5  \\
  \hline
  0 & 28.44  &  11.38  \\
  \hline
  25\% & 28.17 & 10.93 \\
  \hline
  50\%  & {\bf27.91} & {\bf10.82}  \\
  \hline
75\%  & 28.48&  11.07 \\ \hline
  100\% & 28.33 & 10.94 \\
  \hline
\end{tabular}
}{
 \caption{Performance (\%) of model-D by using various percentages of data from the highly noisy subset.}
 \label{tbl:test_model_noisy}
}
\capbtabbox{
\begin{tabular}{c|p{1.3cm}<{\centering}|p{1.3cm}<{\centering}}
  \hline
  Networks & Top1 & Top5  \\
  \hline
Inception\_v2& 27.91  &  10.82  \\
  \hline
  Inception\_v3 & 22.21 & 7.88 \\
  \hline
  Inception\_v4 & 21.97 & 6.64  \\
  \hline
 Inception\_resnet\_v2  & {\bf20.70} &{\bf 6.38}  \\
  \hline
\end{tabular}
}{
 \caption{Performance (\%) of model-D by using various networks.}
 \label{tbl:test_model_networks}
}
\end{floatrow}
\end{table*}

\subsubsection{Comparisons with the state-of-the-art methods.}
Our method is evaluated by comparing it with recent state-of-the-art approaches developed specifically for learning from label noise, such as CleanNet \cite{Lee2017}, FoodNet \cite{pandey2017foodnet} and Patrini \emph{et. al.}'s approach \cite{Patrini2017}. Experiments and comparisons are conducted on four benchmarks: WebVision \cite{li2017webvisiondata}, ImageNet \cite{DengDSLL009}, Clothing1M \cite{xiao2015learning} and Food101 \cite{bossard2014food}.  Model-D with Inception\_v2 is used in all our experiments. By following \cite{Lee2017}, we use the training set of WebVision to train the models, and test on the validation sets of the WebVision and ILSVRC, both of which has the same 1000 categories. On the Clothing1M, we conduct two groups of experiments by following \cite{Lee2017}, we first apply our curriculum-based training method to one million noisy data, and then use 50K clean data to fine-tune the trained model. We compare both results against CleanNet \cite{Lee2017} and the approach of Patrini \emph{et. al.} \cite{Patrini2017}.

Full results are presented in Table \ref{tbl:compar_test}. CurriculumNet improves the performance of our baseline significantly in all four databases. Furthermore, our results compare favorably against recent CleanNet on all datasets, with consistent improvements ranged from about 1.5\% to 3.3\%. Particularly, CurriculumNet reduces Top 5 error of the CleanNet from 12.2\% to 10.8\% on the WebVision data. In addition, CurriculumNet also outperforms Patrini \emph{et. al.}'s approach (19.6\%$\rightarrow$18.5\%) \cite{Patrini2017} on the Clothing1M. On the Food101, CurriculumNet, trained with 20\% additional noise data with \emph{completely random labels}, achieved substantial improvements over both CleanNet (16.0\%$\rightarrow$12.7\%) and FoodNet (27.9\%$\rightarrow$12.7\%) \cite{pandey2017foodnet}. These remarkable improvements confirm the advances of CurriculumNet, demonstrating strong capability for learning from massive ammounts of noisy labels.

\subsubsection{Train with more clean data: WebVision+ImageNet.}
We evaluate the performance of CurriculumNet by increasing the amount of clean data in the training set of WebVision. Since ImageNet data is fully cleaned and manually annotated, a straightforward approach is to simply combine the training sets of WebVision and ImageNet data. We implement CurriculumNet with Inception\_v2 by considering ImageNet data as an additional clean subset, and test the results on the validation sets of both databases. Results are reported in Table \ref{tbl:webvision_imagenet}.

We summarize key observations as follows. (i) By combining WebVision data into ImageNet data, the performance is generally improved due to the increased amount of training data. (ii) Performance of the proposed CurriculumNet is improved significantly on both validation sets by increasing the amount of the clean data (ImageNet), such as 10.8\%$\rightarrow$8.5\% on WebVision, and 15.1\%$\rightarrow$7.1\% on ImageNet. (iii) By using both WebVision and ImageNet as training data, CurriculumNet is able to improve the performance on both validation sets. For example, it reduces the Top 5 error of WebVision from 9.0\% to 8.5\% with a same training set. (iv) On ImageNet, CurriculumNet boosts the performance from a Top 5 error of 8.6\% to 7.1\%, by leveraging additional noisy data (e.g., WebVision). This performance gain is significant on ImageNet, which further confirms the strong capability of CurriculumNet on learning from noisy data.
\begin{table}[tp]
\begin{center}
\caption{Comparisons with most recent results on the Webvision, ImageNet, Clothes-1M and Food101 databases. For the Webvision and ImageNet, the models are trained on WebVision training set and tested on WebVision and ILSVRC validation sets.}
\label{tbl:compar_test}
\vspace{-2mm}
\begin{tabular}{p{2.2cm}|p{2.4cm}<{\centering}|p{2.4cm}<{\centering}|p{2cm}<{\centering}|p{2cm}<{\centering}}
\hline
\multirow{2}{*}{Mothed}& \textbf{ WebVision}& \textbf{ImageNet}& \textbf{Clothing1M}& \textbf{Food101} \\ \cline{2-5}
&Top-1(Top-5)&Top-1(Top-5)&Top-1&Top-1 \\
\hline
Baseline\cite{Lee2017}& 32.2(14.2)& 41.1(20.2)&24.8&18.3 \\
\hline
CleanNet \cite{Lee2017} &29.7(12.2) & 36.6(15.4)&20.1 & -- \\
\hline
MentorNet \cite{jiang2017mentornet} & 29.2(12.0) &37.5(17.0) & -- & -- \\
\hline
\hline
Our Baseline &30.3(13.0) & 37.1(16.4)& 24.2& 15.0 \\
\hline
CurriculumNet & {\bf27.9(10.8) } &{\bf 35.2(15.1)} &{\bf 18.5}& {\bf 12.7} \\
\hline
\end{tabular}

\end{center}
\end{table}

\begin{table}[tp]
\begin{center}
\caption{Performance on the validation sets of ImageNet and WebVision. Models are trained on the training set of ImageNet, WebVision or ImageNet+WebVision. }
\label{tbl:webvision_imagenet}
\vspace{-2mm}
\begin{tabular}{p{6cm}|p{1.5cm}<{\centering}|p{1.2cm}<{\centering}|p{1.2cm}<{\centering}|p{1.2cm}<{\centering}}
  \hline
\multirow{2}{*}{Traning Data}& \multicolumn{2}{c}{\textbf{ WebVision}}&  \multicolumn{2}{c}{\textbf{ImageNet}} \\  \cline{2-5}
&Top-1&Top-5&Top-1&Top-5 \\
\hline
 ImageNet&32.8&13.9 & 26.9&8.6 \\
 \hline
 ImageNet+WebVision& 25.3&9.0& 25.6&7.4 \\
 \hline
\hline
CurriculumNet(WebVision)& 27.9 & 10.8 & 35.2 & 15.1 \\
CurriculumNet(WebVision+ImageNet) & {\bf24.7} &{\bf 8.5} & {\bf24.8}&{\bf 7.1} \\
\hline
\end{tabular}
\end{center}
\end{table}

\section{Conclusion}
We have presented  CurriculumNet - a new training strategy able to
train CNN models more efficiently on large-scale weakly-supervised web images, where no human annotation is provided.
By leveraging the idea of curriculum learning, we propose a novel learning curriculum by measuring data complexity using cluster density. We show by experiments that the proposed approaches have strong capability for dealing with massive, noisy sets of labels. They not only reduce the negative affect of noisy labels, but also, notably, improve the model generalization ability by using the highly noisy data. The proposed CurriculumNet achieved state-of-the-art performance on the Webvision, ImageNet, Clothing-1M and Food-101 benchmarks. With an ensemble of multiple models, it obtained a Top 5 error of 5.2\% on the Webvision Challenge 2017, which outperforms the other submissions by a large margin of about 50\% relative error rate.

\clearpage

\bibliographystyle{splncs04}
\bibliography{egbib}	

\end{document}